\def\year{2019}\relax
\documentclass[letterpaper]{article} 
\usepackage{aaai19}  
\usepackage{times}  
\usepackage{helvet}  
\usepackage{courier}  
\usepackage{url}  
\usepackage{graphicx}  
\usepackage{latexsym}
\usepackage{comment}
\usepackage{url}
\usepackage{amsmath}
\usepackage{amssymb}
\usepackage{algorithm}
\usepackage{adjustbox}
\usepackage{rotating, graphicx}
\usepackage{booktabs}
\usepackage{multirow}
\usepackage{xcolor}
\DeclareMathOperator*{\argmax}{arg\,max}

\usepackage[noend]{algpseudocode}
\begin{document}
 
%
\title{Unsupervised Learning of Interpretable Dialog Models}
\author{
Dhiraj Madan, Dinesh Raghu, Gaurav Pandey and Sachindra Joshi\\
IBM Research AI
}

\maketitle
\begin{abstract}
Recently several deep learning based  models have been proposed for end-to-end learning of dialogs. While these models can be trained from data without the need for any additional annotations, it is hard to interpret them. On the other hand, there exist traditional state based dialog systems, where the states of the dialog are discrete and hence easy to interpret. However these states need to be handcrafted and annotated in the data. To achieve the best of both worlds, we propose Latent State Tracking Network (LSTN) using which we learn an interpretable model in unsupervised manner. The model defines a discrete latent variable at each turn of the conversation which can take a finite set of values.  Since these discrete variables are not present in the training data, we use EM algorithm to train our model in unsupervised manner. In the experiments, we show that LSTN can help achieve interpretability in dialog models without much decrease in performance compared to end-to-end approaches.
\end{abstract}

\section{Introduction}

Recently, there have been several approaches \cite{VL15,Li15,Ser16,Li16,Ser17} proposed for end-to-end learning of dialogs. Most of these approaches have an encoder-decoder architecture. The encoder understands the conversation so far by encoding it as a context vector, while the decoder generates the response based on the context vector. As the context vector is in continuous space, it is hard to interpret what the system has understood. Moreover, it is hard to interpret why a particular response was generated. More importantly, the model provides no means to control the type of responses the system can generate. In spite of being a black box, these approaches have gained popularity as they can easily adapt to new domain and do not require additional annotations on data.

On the other end of the spectrum are the traditional dialog systems. They cannot easily adapt to new domains as they require additional annotations on the data. However they are well interpretable and provide complete control over the system, as they demand a discrete state space to be defined for what the system can understand - \textit{belief state} and what it can respond with - \textit{action state}. At each turn of the conversation, the belief state is updated based on the user input and the previous belief state. The belief state is mapped to the action state, based on which a response is generated. These \textit{state-based} dialog systems are usually designed as Markov decision processes or partially observable Markov decision processes. 
These states provide control over dialog systems and also help in interpreting its behavior. Human-intervention is necessary to define these states and annotate each dialog in the data that makes it hard to scale to new domains.

Recently, there has been a push towards reducing the amount of human intervention in state-based dialog systems \cite{Wen17} without compromising on interpretability. \citeauthor{Wen17} proposed a deep learning based approach that learns the action space of a state-based dialog system in an unsupervised manner. However, the approach still requires the belief state to be hand-crafted and annotated for each turn in a dialog. 
There are also some efforts~\cite{ZLE18} for making end-to-end dialog systems more interpretable.  Zhao et al. proposed a modification to end-to-end models where in they augment the context vector with a discrete valued vector. This helps in partially understanding why the response was generated and also provide control over the generated responses. However, as the belief state is still in a continuous space, interpreting the system's understanding is still an open problem. Thus we see that the two strands of works are steadily moving towards the common goal of building a fully interpretable dialog model without the need for human intervention. 

In this paper, we propose an approach for unsupervised learning of fully interpretable dialog models. 
We propose a Latent State Tracking Network (LSTN) to learn internal discrete states in an unsupervised manner. The network encodes the conversation-so-far into a discrete latent state using a transition model, while the emission model generates a response based on the encoded state.  
Since the proposed model is unsupervised, the discrete states are not available during training. Hence we propose an expectation-maximization (EM) based solution for jointly learning the states as well as the transition and emission models. Once the model has been trained, we can infer the state associated with a new user utterance using the transition model. Furthermore, we can generate the response that corresponds to the state using the emission model. 

\noindent To summarize, we make the following contributions:
\begin{enumerate}
\item We define a framework for learning interpretable dialog models in an unsupervised manner.
\item We propose a novel Latent State Tracking Network (LSTN) for learning interpretable dialog models from conversations without any supervision.
\item We propose an EM-algorithm for jointly learning the latent states as well as the transition and emission modules in an LSTN. 
\item We also show that in this process of discretization we do not lose much over the state-of-the-art, deep learning models for dialog, but gain in terms of having an interpretable model which can be easily modified using domain knowledge.
\end{enumerate}

\begin{figure*}[h]
\centering
 \includegraphics[width=.97\textwidth]{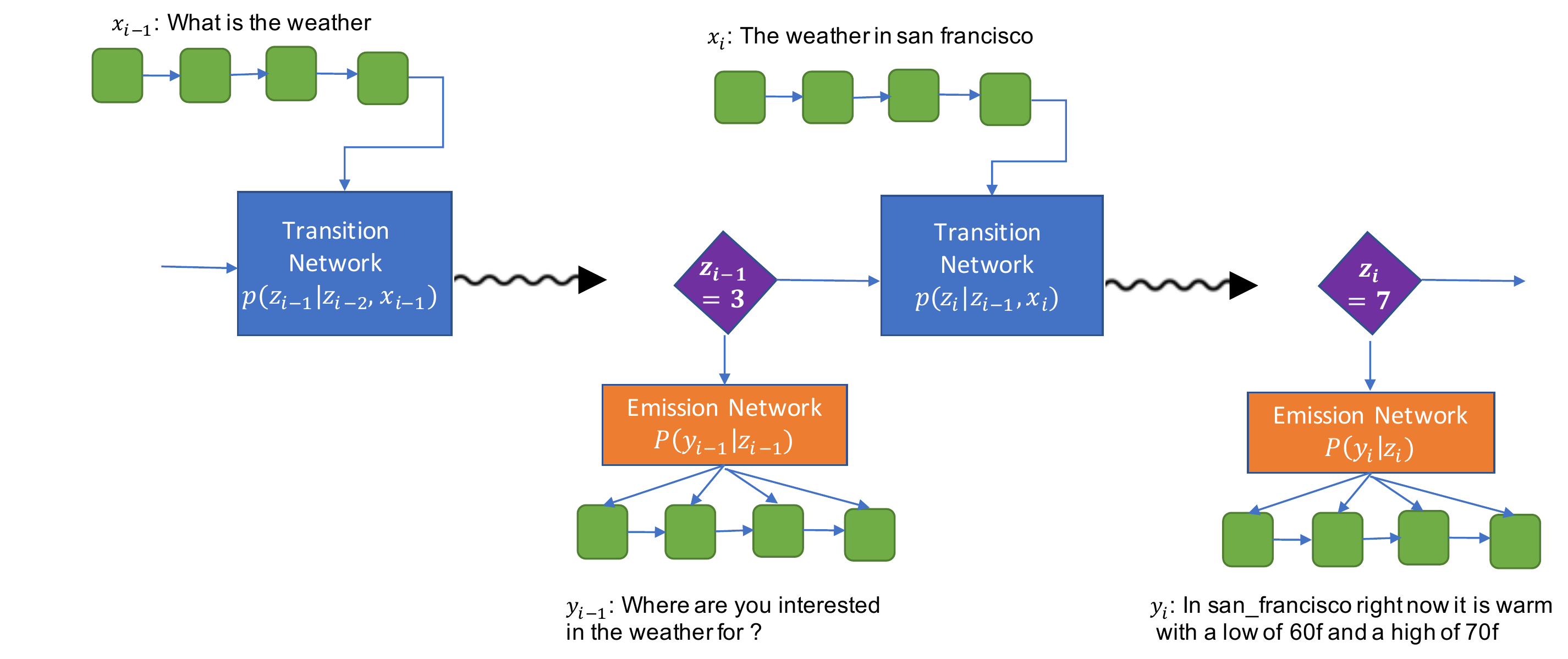}
    \caption{The Latent State Tracking Network for two steps of a conversation.}
    \label{fig:dl_network}
\end{figure*}




\section{Latent State Tracking Network}
\subsection{The Proposed Model}
Let a dialogue $D = \{x_1,y_1,\ldots,x_N,y_N\}$ be represented as a sequence of utterances where $x_i$ and $y_i$ are the user utterance and agent response at the $i^{th}$ turn. Given a set of such dialogues, we wish to learn a interpretable dialog model $\mathbb{M}$ which encodes the conversation so far using a discrete state variable and then samples a value from the state variable to generate an agent response $y_i$.

The state variable $z_i \in \{1,..,K\}$ at any turn $i$, is computed using the user utterance $x_i$ at turn $i$, along with the previous turn's state variable $z_{i-1}$. We model this dependency using the \textit{transition distribution} $p(z_i|z_{i-1}, x_i)$. The agent response $y_i$ is then generated based on the state variable $z_i$. The relation between the response and the discrete state variable is modeled using an emission distribution $p(y_i|z_i)$. An illustration of the flow of dialog using our latent state tracking network is shown in Figure~\ref{fig:dl_network}.

\label{sec:approach}
\begin{figure}[h]
\centering
 \includegraphics[width=0.3\textwidth]{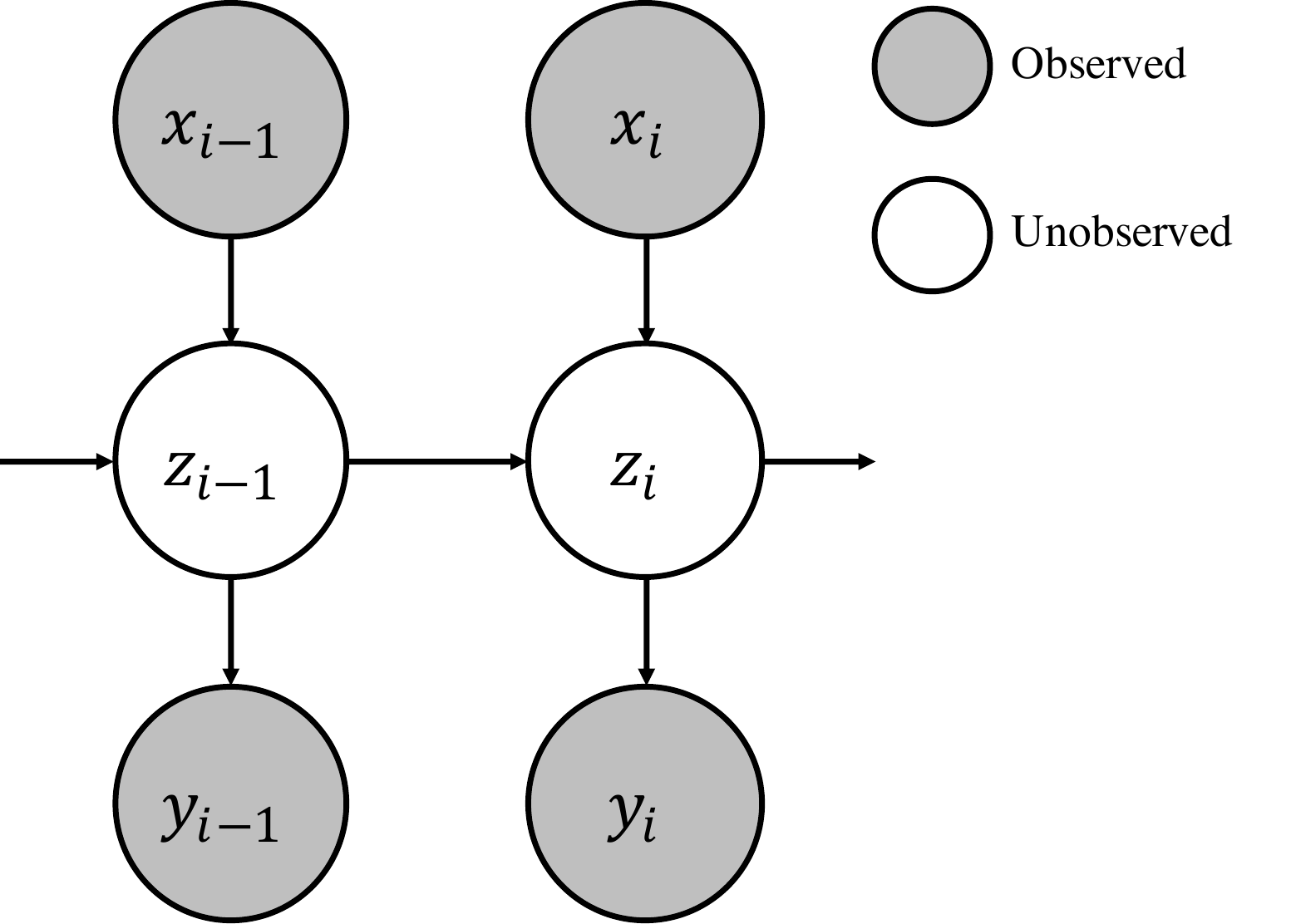}
    \caption{Plate notation of the Latent State Tracking Network}
    \label{fig:graphical}
\end{figure}

A graphical model representation of LSTN is given in Figure~\ref{fig:graphical}.
The joint distribution of the agent responses  $\mathbf{y} = (y_1, \ldots,  y_N)$ and the belief states $\mathbf{z} = (z_1, \ldots, z_N)$ given the user utterances $\mathbf{x} = (x_1, \ldots, x_N)$ for a given conversation can be written as:
\begin{equation}
\begin{aligned}
p(\mathbf{z},\mathbf{y} \vert \mathbf{x}) 
=\prod_{i=1}^N p(z_i \vert z_{i-1},x_i) p(y_i \vert z_i)
\end{aligned}
\end{equation}

\noindent Note that there are two key distributions in this model:
\begin{enumerate}
\item The transition distribution which models the probability of moving to a new state $z_i$ given the previous state $z_{i-1}$ and current user utterance $x_i$.
\item The emission distribution which models the probability of generating response $y_i$ given the current state $z_i$.
\end{enumerate}

In order to completely define the model, we need to explain the computation of the above distributions from the utterances in a conversation.

\subsubsection{The Transition Distribution:}
Here, we need to model the probability of observing a new state $z_i$ given the previous state $z_{i-1}$ and the user utterance $x_i$. We use an LSTM network to embed the user utterance to a hidden state representation $h(x_i)$. For modeling the transition distribution, the states $\{1, \ldots, K\}$ are represented using continuous vectors $\{v_1, \ldots, v_K\}$. Hence, for the state $z_{i-1}$, we fetch the corresponding vector representation $v_{z_{i-1}}$. This vector is then concatenated with the hidden state representation of the utterance and then fed to a classifier with softmax outputs.  The classifier outputs a probability distribution over the next states. Hence, the probability of the next state $z_i$ given the previous state $z_{i-1}$ and the user utterance $x_i$ is given by
\begin{equation}
p(z_i|z_{i-1}, x_i) = \text{softmax}(W[h(x_i); v_{z_{i-1}}] + b) \,,
\end{equation}
where $W, b$, the network $h$ and the embeddings $v_{z}$ are parameters that are learnt during training.

\subsubsection{The Emission Distribution:} 
Given the current state $z_i$, this distribution models probability of all possible responses. To model this distribution, the states $\{1, \ldots, K\}$ are represented using continuous vectors $\{r_1, \ldots, r_K\}$. We feed the embedding of the current state to the decoder LSTM which outputs a sequence of distributions over the words. The probability of a response $y_i = (w_1, \ldots, w_M)$ conditioned on the state $z_i$ is given by
\begin{equation}
    p(y_i|z_i) = \prod_{j=1}^M p(w_j|w_1, \ldots, w_{j-1},z_i)
\end{equation}

\begin{algorithm*}[]
\caption{Training Algorithm}\label{algo1}
\begin{algorithmic}[1]
\Procedure{computeCost} 
{}
\State {\textbf{Input:} Dialog Utterances $\{(x^{(i)},y^{(i)})\}_{i=1}^{N_i}$,  Parameter Weights $\Theta$, Posterior $q(z_i \vert z_{i-1}, \mathbf{x}, \mathbf{y})$}
\State{\textbf{Output:} Log likelihood $L(\Theta)$}
\State{Compute $f_N(\Theta, z_{N-1})$ using the transition and emission distribution as defined in~\eqref{eq:fN}}
\For{{$i \leftarrow N-1$ downto $1$ }}
        \State{Compute $f_i(\Theta, z_{i-1})$ from $f_{i+1}(\Theta, z_{i})$ using the transition and emission distribution as defined in~\eqref{eq:M_rec}}.
      \EndFor
\State \textbf{return} $f_0(\Theta,z_0=0)$
\EndProcedure
\end{algorithmic}
\end{algorithm*}

\subsection{Training the LSTN}
In order to train the model, we need to maximize the marginal log-likelihood of the responses given the user utterances. Hence, we need to marginalize out the states $\mathbf{z} = (z_1, \ldots, z_N)$ from the model. The corresponding marginal log-likelihood for a single conversation is given by
\begin{align}
L(\Theta) &= \ln(p( \mathbf{y} \vert \mathbf{x}, \Theta))\\
&= \ln\left(\sum_{\mathbf{z}} \prod_i p(z_i \vert z_{i-1},x_i;\Theta) p(y_i \vert z_i ; \Theta)\right).
\end{align}

In order to simplify the computation of the above quantity, we lower-bound it using an EM algorithm. In particular, for any distribution $q$ over the states $(z_1, \ldots, z_N)$, the above quantity can be rewritten as
\[L(\Theta)= \mathbb{E}_{\mathbf{z}}\left[\ln\left(\frac{p(\mathbf{y}, \mathbf{z}\vert \mathbf{x},\Theta)}{q(\mathbf{z})}\right)\right] +\mathrm{KL} \left(q \| p(\mathbf{z}\vert \mathbf{y}, \mathbf{x} ,\Theta)\right) \,,\]
where the expectation is over the distribution $q(\mathbf{z})$.
Since, $\mathrm{KL}$ divergence is always non-negative, the first term in the above equation is a lower bound to the log-likelihood for any choice of $q$. Furthermore, this lower bound is tight, when $q$ exactly equals the posterior distribution over the states given all the user utterances and the agent responses in the conversation .

Hence, the training proceeds as follows. In the first step, also referred to as the E-step in literature, we compute the posterior distribution over all the states of a given conversation based on our current estimate of the parameters. In the M-step, we maximize the expectation of the joint log-likelihood with respect to the posterior obtained in the E-step. We discuss these steps in further detail below.

\subsubsection{The E-step:}
As discussed in the previous section, the prior distribution over the states of an LSTN given the user utterances factorizes as follows:
\begin{equation}
p(\mathbf{z}|\mathbf{x}) = \prod_{i=1}^N p(z_i|z_{i-1}, x_i) \,,
\end{equation}
where $z_0=0$ is the default state at the beginning of a conversation. Here, we will discuss the computation of the posterior distribution over the states given the user utterances and the agent responses. As with the prior, the posterior distribution over the states factorizes. That is,
\begin{equation}
\begin{aligned}
p(\mathbf{z}|\mathbf{x}, \mathbf{y}) = \prod_{i=1}^N p(z_i|z_{i-1}, \mathbf{x}, \mathbf{y}) 
\end{aligned}
\end{equation}
For the sake of brevity, we refer to $y_i, \ldots, y_N$ as $\mathbf{y}_{i:N}$. The same notation is used for sequence of user utterances and latent states.
In order to compute the posterior, we note that given the previous state, the next state is independent of all previous agent responses. That is:
\begin{align*}
p(z_i \vert z_{i-1},\mathbf{y},\mathbf{x})&=p(z_i \vert z_{i-1}, y_{i:n},\mathbf{x})\\
&\propto p(z_i, y_{i:n} \vert z_{i-1},\mathbf{x} )\\
\end{align*}
To compute the above distribution, we use dynamic programming. In particular, the above distribution can be expressed in terms of the corresponding distribution at timestep $i+1$ as follows: 
\begin{equation}
\label{eq:E_step}
\begin{aligned}
 p&(z_i, y_{i:N} \vert z_{i-1}, \mathbf{x} ) \\
 &= p(z_i \vert z_{i-1},x_i) p(y_i\vert z_i)  \sum_{z_{i+1}} p(z_{i+1}, y_{i+1:N} \vert z_i, \mathbf{x} )
 \end{aligned}
\end{equation}
Note that the distribution within the summation has the same form as the distribution that we wish to compute. Hence, the desired distribution at timestep $i$ can be computed recursively from the corresponding distribution at timestep $i+1$. Moreover, the distribution at the last timestep can be computed directly as follows:
\begin{equation}
    p(z_N, y_N|z_{N-1},\mathbf{x}) = p(z_N|z_{N-1}, x_N)p(y_N|z_N)
\end{equation}
Thus, we can run the above computation over the $N$ turns of the conversation to obtain the posterior distribution of each latent state.
\subsubsection{The M-step:}
Having obtained the posterior, we use it for maximizing the expected complete log-likelihood of the agent responses and the latent states. In particular, we need to maximize
\begin{equation}
     \mathbb{E}_{\mathbf{z} \sim p(\mathbf{z}\vert \mathbf{y}, \mathbf{x}, \Theta^{old})} \ln p(\mathbf{z},\mathbf{y} \vert \mathbf{x}, \Theta) 
\end{equation}
Here, $\Theta_{old}$ in the posterior refers to the fact that the posterior has been evaluated using the current parameters and will be held fixed during the M-step. The above expectation is computed recursively using a Viterbi based approach. 
We equate the expectation of the log-likelihood of the last $N-i$ states and agent responses to $f_i(\Theta, z_{i-1})$. That is:
\begin{equation}
    \begin{aligned}
    f_i(\Theta, z_{i-1}) = \mathbb{E}_{\mathbf{z}_{i:N}}  \ln p(\mathbf{z}_{i:N}, \mathbf{y}_{i:N}| \mathbf{x}_{i:N}, z_{i-1}, \Theta)\,,
    \end{aligned}
\end{equation}
where the expectation is over the posterior distribution of the latent states. Note that the objective that we wish to optimize is $f_1(\Theta, z_0)$, where $z_0$ is the default start state. To compute this quantity, we note that $f_i(\Theta, z_{i-1})$ can be expressed as function of $f_{i+1}(\Theta, z_{i})$ as follows:
\begin{equation}
\label{eq:M_rec}
\begin{aligned}
    f_i & (\Theta, z_{i-1})= \mathbb{E}_{z_{i}} \left[f_{i+1}(\Theta, z_{i}) \right.\\
    &+ \left. \ln(p(z_i \vert z_{i-1},x_i, \Theta))+ \ln(p(y_i \vert z_i, \Theta))\right] \,,
    \end{aligned}
\end{equation}
where the expectation is over the posterior distribution of $Z_i$.
Finally, we note that $f_N(\Theta, z_{n-1})$ can be computed directly to begin the recursion as follows:
\begin{equation}\label{eq:fN}
\begin{aligned}
    f_N  (\Theta, z_{N-1}) =& \mathbb{E}_{Z_{N}} \left[ \ln(p(z_N \vert z_{N-1},x_N, \Theta))\right.\\
    &+ \left. \ln(p(y_N \vert z_N, \Theta))\right] 
    \end{aligned}
\end{equation}

The computation of $f_1(\Theta, z_0)$ from $f_N(\Theta, z_N)$ constitutes the forward pass of the M-step and is listed in Algorithm~\ref{algo1}. Note that each step of the computation is differentiable, and hence, the objective is a differentiable function of the transition and emission distributions. Hence, during the backward pass, we backpropagate the gradient all the way from the final objective to the transition and emission distributions.

\subsection {Inference with the given model}
\label{sec:inference}
In this section we will discuss how the trained model is used for generating the  response utterance given the context consisting of previous user utterances and agent responses . 
There are two parts to our inference:-

\subsubsection{Emission Module:} Here given a dialog state, we need to generate the mostly likely responses associated with the same. Having learnt the distribution $p(y|z)$, as a decoder RNN, we use this to generate top responses for each value of $z$. For each value of $z$ from $1$ to $K$, we initialize the hidden state of decoder RNN with vector $r_z$ and perform beam search to generate the top responses. In our experiments we used a beam size of 10. This step is performed only once and is not repeated for new test examples.
At test time it will suffice to use the top responses associated with a state or sample one from the top 10 generated through beam search. 

\subsubsection{Transition Module:} This module computes a distribution over current state given the past state and the new user utterance. During inference, we use this module to obtain the distribution of each state given the past user utterances. In particular, the distribution of the $i^{th}$ state given all the user utterances till step $i$ can be expressed as follows:

\[  p(z_i \vert \mathbf{x}_{1:i}) = \sum_{z_{i-1}} p(z_i \vert z_{i-1},\mathbf{x}_i) p(z_{i-1} \vert \mathbf{x}_{1:i-1}) 
\]
During inference, we can generate the response corresponding to the most probable hidden state.
i.e. we compute $\bar{z}_i= \argmax_{z_i} p(z_i \vert \mathbf{x}_{1:i})$. We then produce the most likely response corresponding to $\bar{z}_i$ using emission module.

\section{Experimental Setup} \label{sec:experiments}
\subsection{Datasets}
We perform experiments on four dialog datasets: Stanford Multi-Domain Dataset (SMD) \cite{smd}, CamRest \cite{camrest}, DSTC6 \cite{dstc6}, and Car Assistant Dialog Dataset. CamRest and SMD were collected through Amazon Mechanical Turk using Wizard-of-Oz framework. DSTC6 dataset is a corpus of (context, response) pairs rather than entire dialogs. As complete dialogs were required to train our system, we filtered\footnote{dialogs with at least one restaurant suggestion were retained} the pairs that constituted complete dialogs. Since the dataset was synthetically generated using a set of templates, it was fairly simple to filter them.

Task oriented datasets are usually grounded to a knowledge base. CamRest, SMD and DSTC6 are all task oriented dialogs. Modeling the interaction with the KB is a crucial part of learning task oriented dialogs. Since our focus is to evaluate the system on interpretability, we removed the dependency on the knowledge base by anonymizing each KB entity present in the dialogs. For example, the utterance ``Let's go with \textit{Japanese} food, I will keep \textit{Korean} for next time" will be anonymized as ``Let's go with \textit{cuisine\_0} food, I will keep \textit{cuisine\_1} for next time". The anonymized datasets were used for all experiments.
We have also learnt our model on each task (i.e. scheduling, navigation and weather related queries) of SMD dataset separately . 
The Car Assistant Dialog Dataset (CADD) is a set of 986 conversations between an in-house\footnote{details to be added in the camera ready version} car assistant bot and its users. The bot is designed to help with navigation and controlling various devices in the car. Some statistics of all the datasets used are summarized in Table \ref{table:datasets}.

\begin{table}[h!]
\centering
\begin{tabular}{l c c c c }
\toprule
\multirow{2}{*}{\textbf{Dataset}} &  \multicolumn{3}{c}{\textbf{No. of Dialogs}} & \textbf{Avg. No.}  \\                                     
& \textbf{Train}  &  \textbf{Dev}  & \textbf{Test} & \textbf{of Turns} \\ 
\midrule
SMD (Nav.) & 800 & 100 & 100 & 3.28 \\ 
SMD (Wea.) & 797 & 99 & 100 & 2.69 \\
SMD (Cal.) & 828 & 103 & 104 & 1.86 \\
SMD (All) & 2425 & 302 & 304 & 2.6 \\
DSTC6 & 1661 & 185 & 1000 & 18.70\\ 
CADD & 786 & 100 & 100 & 4.73\\
CamRest & 406 & 135 & 135 & 4.06 \\

\bottomrule
\end{tabular}
\caption{Statistics of various datasets used}
\label{table:datasets}
\end{table}


\subsection{Training}
Adam optimizer was used for training \cite{kin14}. The hyperparameters were selected based on perplexity on a held-out validation set. The learning rate was sampled from the set $\{0.01,0.001,0.0001\}$, dimension of word embeddings from $\{16,32,64\}$, the number of distinct latent states $K$ from $\{8,16,32,64,128\}$. We experimented with having same versus different embeddings for the latent states  while computing transition and emission distributions. 


 
\section{Experimental Results}

\subsection{Recoverability}
As LSTN generates agent response only based on the latent discrete state, we wish to quantify how well the responses can be generated, given we know the optimal state for each response. The optimal state $\bar{z}$ of a response $y$ is computed as  $\bar{z}=\argmax_{z} p(y|z)$. Using $\bar{z}$, we then generate the most likely response $\bar{y}$ as $\bar{y}=\argmax_{y'} p(y'\vert \bar{z})$. The recoverability score of a response $y$  is defined as the BLEU score \cite{Pap02} between the original response $y$ and generated response $\bar{y}$. The recoverability score of a dataset is then defined as the average of the recoverability scores of all the test responses. Table \ref{table:rec} lists the recoverability scores of various datasets. We observe that recoverability scores are much higher for CADD and DSTC6 datasets, as these datasets contain templatized machine generated responses. 



Recoverabilty score also defines the upper bound for the LSTN model for a given $K$. In particular, if we are able to learn a perfect state transition distribution, we can achieve a test BLEU score equal to the recoverabilty score. 
In Table \ref{table:rec}, we also show the test BLEU score obtained by a non-interpretable HRED \cite{Ser16} model that uses real valued context state vectors. We observe the recoverability score are higher than the BLEU scores achieved by HRED. This shows that if we can learn a perfect state transition model, we can perform as good or better than non-interpretable models like HRED.

\begin{table}[h!]
\centering
\begin{tabular}{l c c}
\toprule
\multirow{2}{*}{\textbf{Dataset}}  & \textbf{Recoverability}  & \textbf{HRED}\\
& \textbf{Score} & \textbf{BLEU} \\
\midrule
CADD & 80.2    & 66.2   \\ 
DSTC6                         & 91.9   &  88.9   \\ 
SMD (All) & 14.67 & 14.00\\
SMD (Cal.)                & 18.7 & 17.9      \\ 
SMD (Nav.)              & 13.4  & 9.3      \\ 
SMD (Wea.)                 & 21.3  & 15.2    \\
CamRest                       & 14.9 & 12.1     \\
\bottomrule
\end{tabular}
\caption{Recoverability scores of LSTN with $K=32$ and HRED BLEU scores on various datasets}
\label{table:rec}
\end{table}

\subsection{End-to-End Evaluation}


To illustrate the advantage of jointly modeling the transitions and emissions (as in LSTN), we compare it against a model that learns the emissions and the transitions in a pipelined fashion. We call the latter model as \textit{split-LSTN}. This model is learnt in two phases. In the first phase, the conversation-so-far is encoded using an LSTM and mapped to a discrete-value from which the response is generated. The likelihood of the response is maximized using EM algorithm to learn the latent states and the emissions. In the second phase, we learn the transitions between the latent states obtained in the previous phase.  Table \ref{table:bleu} lists the BLEU scores obtained on various datasets using LSTN and split-LSTN. We can observe that LSTN consistently outperforms split-LSTN on all datasets. Hence we conclude jointly modelling transitions and emissions results in better performance.

\begin{table}[h!]
\centering
\begin{tabular}{l c c}
\toprule
\multirow{2}{*}{\textbf{Dataset}} & \multicolumn{2}{c}{\textbf{BLEU Score}} \\
\cmidrule{2-3}
& \textbf{Split-LSTN} & \textbf{LSTN} \\
\midrule
CADD       & 55.31 & 57.19 \\              
DSTC-6     & 64.07 & 74.1  \\
SMD (Cal.) & 13.04 & 15.53 \\
SMD (Nav.) & 5.86  & 6.49  \\
SMD (Wea.) & 12.36 & 13.35 \\
SMD (All)  & 9.72  &10.57  \\
CamRest    & 9.61  & 10.40 \\
\bottomrule
\end{tabular}
\caption{Comparison of BLEU Scores of responses generated from LSTN and split-LSTN}
\label{table:bleu}
\end{table}

\subsection{Variations with Number of Latent States}

\begin{figure}[h]
\centering
 \includegraphics[width=0.5\textwidth]{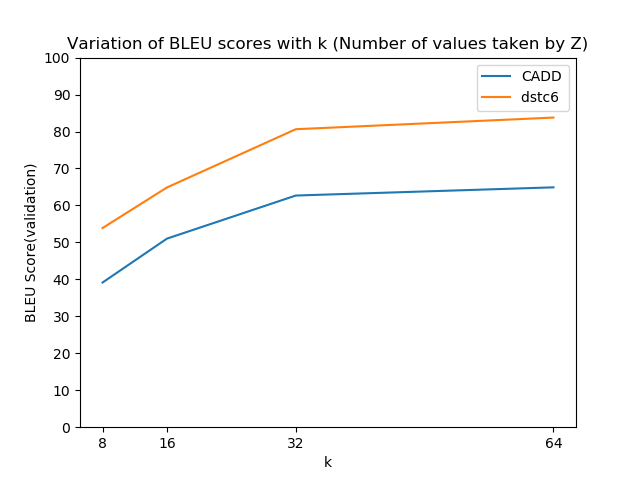}
    \caption{Variation of BLEU Score with number of latent states K }
    \label{fig:vary_k}
\end{figure}

To evaluate the effect of the number of latent states $K$, we trained LSTN for several values for $K$. Figure \ref{fig:vary_k} shows the performance in terms of BLEU score for two datasets. We observe that as the value of $K$ increases the performance also increases and saturates after a while.


\subsection{Qualitative Results}
In this section, we qualitatively evaluate the transition and emission modules of the LSTN model. In order to evaluate the emission module, we list the top responses associated with some of the states in Table~\ref{table:responses_smd_all}. As can be observed, the responses that provide the date/time of an event are grouped together in state $14$. Similarly, state $20$ contains the responses that discuss the weather at a location.
Hence, we conclude the LSTN learns to group similar responses to a single state.

To evaluate the transition module, we simulated our inference algorithm on training data, and for each pair of latent states $z, z'$ we look for utterances which lead us from state $z$ to $z'$. We classify each such group of utterances, as an \textit{Intent Class}. We show some of these Intent Classes in Table \ref{table:intents}. As can be observed, the user utterances that cause a transition from state $0$ to state $14$, inquire the date/time of events. Similarly, all the utterances that cause a transition from state $0$ to state $20$, ask for the weather at particular locations. Hence, it can be concluded that the LSTN learns to capture user utterances with similar intents together.

\begin{figure}[h]
\centering
 \includegraphics[width=0.45\textwidth]{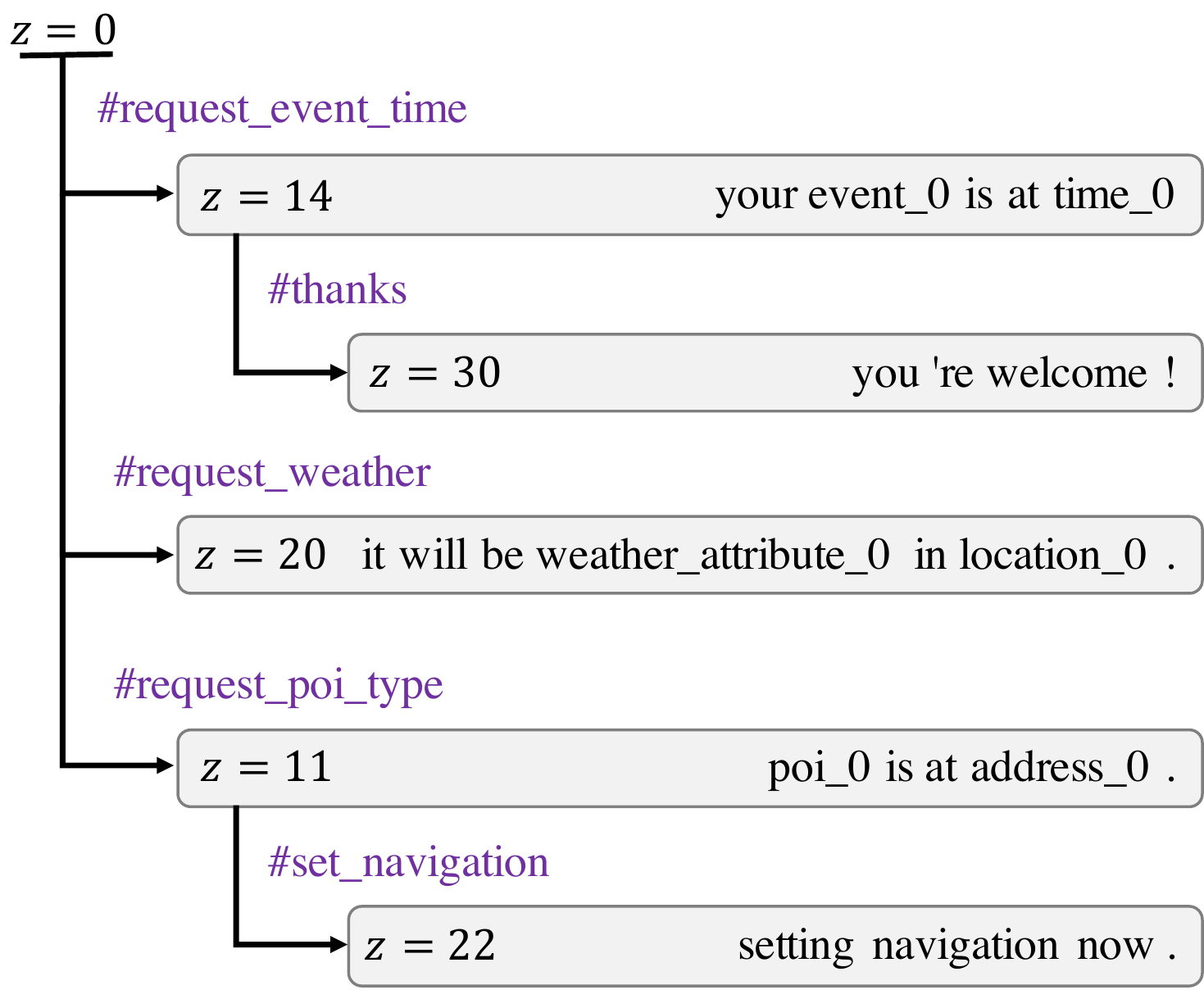}
    \caption{The transition and emission represented to provide better intuition }
    \label{fig:dialog-flow}
\end{figure}

Finally, we combine the power of transition and emission distributions in the model to learn a single dialog tree for conversations. In Figure \ref{fig:dialog-flow} we illustrate a part of the dialog tree for SMD dataset. The nodes of this tree indicate the possible states of the dialog system. These correspond to the values taken by the latent variable in our model and their top responses.  
The edges of this tree indicate the possible transitions between the states. 
These kind of trees can be modified by domain expert and can be used in standard dialog frameworks such as Google Dialog Flow\footnote{https://dialogflow.com/}, IBM Watson Assistant\footnote{https://www.ibm.com/watson/ai-assistant/} and Microsoft Bot Framework\footnote{https://dev.botframework.com/}. 

\begin{table*}[h!]
\centering
\begin{tabular}{c l l l }
\toprule
\textbf{Latent State} & \textbf{Response at Rank 1} & \textbf{Response at Rank 2}  & \textbf{Response at Rank 3} \\ \midrule
$z=14$ & your \textit{event\_0} is at \textit{time\_0} & \begin{tabular}[c]{@{}l@{}}your \textit{event\_0} is on \textit{date\_0} at \textit{time\_0}, \\ drive, carefully !\end{tabular}                       & \begin{tabular}[c]{@{}l@{}}your event\_0 is date\_0 at time\_0\end{tabular}            \\ \midrule
$z=20$                                                                  &\begin{tabular}[c]{@{}l@{}} it will be\\ weather\_attribute\_0 in location\_0 \end{tabular} & \begin{tabular}[c]{@{}l@{}} it will not be weather\_attribute\_0 in \\ location\_0 weather\_time\_0 .  \end{tabular}& \begin{tabular}[c]{@{}l@{}} it will not be\\ weather\_attribute\_0 in location\_0 . \end{tabular}\\ \midrule
$z=11$ & poi\_0 is at address\_0 & poi\_0 is distance\_0 away . & poi\_0 is located at address\_0\\\midrule
$z=22$ &setting navigation now & \begin{tabular}[c]{@{}l@{}} it will not be weather\_attribute\_0 \\in location\_0\end{tabular}  & \begin{tabular}[c]{@{}l@{}} it will be weather\_attribute\_0\\ in location\_0 .  \end{tabular}\\\midrule
$z=30$ & you 're welcome !  & you are welcome  & you 're welcome .
\\ \bottomrule
\end{tabular}
\caption{Top response for latent variables in SMD dataset}
\label{table:responses_smd_all}
\end{table*}

\begin{table*}[h!]
\centering
\begin{tabular}{c | c | l | c}
\toprule
\textbf{Previous State} & \textbf{Current State } & \textbf{Train User Utterances}  &  \textbf{Intent Class}\\
\midrule
\multirow{3}{*}{$z_{i-1}= 0$} & \multirow{3}{*}{$z_{i}=14$} & when is my next \textit{event\_0} & \multirow{3}{*}{\#request\_event\_time} \\
& & can you check the time and date of my \textit{event\_0} ?  & \\
& & what time is my \textit{event\_0} scheduled & \\
\midrule
\multirow{3}{*}{$z_{i-1}= 0$} & \multirow{3}{*}{$z_{i}=20$} & on tuesday in \textit{location\_0} find out if it will be \textit{weather\_attribute\_0}  & \multirow{3}{*}{\#request\_weather} \\
& & will it be \textit{weather\_attribute\_0} in \textit{location\_0} on wednesday ? & \\
& & is it \textit{weather\_attribute\_0} in \textit{location\_0} now ? & \\
\midrule
\multirow{3}{*}{$z_{i-1}= 0$} & \multirow{3}{*}{$z_{i}=11$} & get me the address of a \textit{poi\_type\_0} around this area .  & \multirow{3}{*}{\#request\_poi\_type} \\
& & what \textit{poi\_type\_0} s are around ?  & \\
& & car i need the address of a \textit{poi\_type\_0} near me , please help me ! & \\
\midrule
\multirow{3}{*}{$z_{i-1}= 14$} & \multirow{3}{*}{$z_{i}=30$} & great , thanks .  & \multirow{3}{*}{\#thanks} \\
& & thanks  & \\
& & that will do just fine , thanks . & \\
\midrule
\multirow{3}{*}{$z_{i-1}= 11$} & \multirow{3}{*}{$z_{i}=22$} & if that 's the best option , set the gps please .  & \multirow{3}{*}{\#set\_navigation} \\
& & set the gps for there please .  & \\
& & can you set navigation and get us there ?  & \\
 \bottomrule
\end{tabular}
\caption{The transitions between the states of an LSTN. The third column of the above table lists the user utterances that result in state transitions from the states in first column to the ones in the second column. An intuitive label is associated with these utterances in the fourth column. }
\label{table:intents}
\end{table*}

\subsection{Error Ananlysis}
We now brief the two major issue encountered due to quantization of dialogs using discrete latent states.

\subsubsection{Inability to Capture Subtle Variations} Table \ref{table:responses_smd_all} shows the most probable responses generated from a single latent value using beam search. The rank 1 and rank 2 responses corresponding to $z=14$ show subtle variations in the information provided back to the user. One provides just the \textit{time} whereas the other provides both \textit{time} and \textit{date}. Since the top response is always picked for a given $z$, the system would provide just the \textit{time} when both \textit{time} and \textit{date} are requested by the user.

\subsubsection{Duplicates} High probable responses such as ``you are welcome'' were being generated from more than one latent state. Similar trends were observed even in transitions. This issue could also be one reason why LSTN is unable to capture the subtle variations in the responses.

\section{Related Work}
To the best of our knowledge, our work is the first to propose an unsupervised approach for learning interpretable dialog models. The word unsupervised indicates that the dialogs used to train the model are not annotated with any additional labels. In this section, we provide a brief overview of works related to (1) unsupervised learning of dialog, (2) interpretable dialog models  and (3) use of discrete latent states in deep learning. 

\subsubsection{Unsupervised Learning of Dialogs}
Early approaches for learning dialogs from chat transcripts were inspired from machine translation \cite{ritter2011data} and language modelling \cite{Sor15}. \citeauthor{VL15} (\citeyear{VL15}) proposed a deep learning approach based on the sequence-to-sequence model \cite{Sut14}. \citeauthor{Ser16} (\citeyear{Ser16}) proposed an approach that leverages the hierarchical structure of the dialog to model them better. \citeauthor{Ser17} (\citeyear{Ser17}) extended the previous approach by modelling the stochasticity in responses. Even though these models can be trained in an unsupervised manner, they are not interpretable.



\subsubsection{Interpretable Dialog Models} Traditional task oriented dialog systems were built using reinforcement learning (RL) approaches. The systems were either modeled as Markov decision processes \cite{levin2000stochastic,walker2003trainable} or partially observable Markov decision processes \cite{Wil07,gasic2013pomdp}. Recently, there has been efforts \cite{wen2017network,Wen17} to solve RL based approaches using deep learning. These RL based approaches provide interpretability due to their discrete intermediate variable such as states and actions. But the interpretability comes at a cost of handcrafting states, actions and rewards for training the RL model. In our approach, we provide interpretability by learning discrete intermediate states latently without the need for any handcrafting.

\subsubsection{Discrete Latent Variables:} 
An emerging area in deep learning research is to use a set of discrete latent variables in the deep learning model. This was first proposed by \cite{JGP16} and \cite{MMT16}. They introduced Gumbel-Softmax (or Concrete Distribution) to enable reparameterization with discrete/categorical variables.
Later \cite{Oor17} introduced the Vector Quantized-Variational AutoEncoder.

Learning dialogs by augmenting the real-valued context vector with discrete latent variables \cite{Wen17,Wil92} adds a notion of interpretability. Our work differs from these approaches as we only use discrete latent state to capture the entire conversation context. The discrete states helps us in making the entire model easy to interpret and modify. \cite{Wen17} learnt the discrete variables in a semi supervised fashion using variational lower bound and REINFORCE idea \cite{Wil92} to back propagate through sampling step. \cite{ZLE18} have also considered a similar approach, using repramaterization trick through Gumbel-Softmax  \cite{JGP16} instead of REINFORCE. Using these variational approximation techniques to learn the discrete variables would make the training of LSTN harder as noise would get cascaded at each step. Hence we compute the exact posterior using an EM based approach. 




\section{Conclusions}
\label{sec:future}
In this paper, we introduce a novel problem of learning interpretable dialog models in an unsupervised manner. We propose a novel model, Latent State Tracking Network (LSTN) for this task. LSTN learns the discrete latent states using a EM based algorithm. We show that (1) even after discretization the states learnt by LSTN are as good as uninterpretable models (such as HRED) (2) joint learning of emissions and transitions is better than learning them in a pipelined manner and (3) the learnt emissions and transitions are interpretable and meaningful. 



\bibliography{aaai}

\begin{thebibliography}{}

\bibitem[\protect\citeauthoryear{Boureau, Bordes, and Perez}{2017}]{dstc6}
Boureau, Y.-L.; Bordes, A.; and Perez, J.
\newblock 2017.
\newblock Dialog state tracking challenge 6 end-to-end goal-oriented dialog
  track.
\newblock In {\em The 6th Dialog System Technology Challenge}.
\newblock Long Beach, USA: ISCA.

\bibitem[\protect\citeauthoryear{Eric \bgroup et al\mbox.\egroup }{2017}]{smd}
Eric, M.; Krishnan, L.; Charette, F.; and Manning, C.~D.
\newblock 2017.
\newblock Key-value retrieval networks for task-oriented dialogue.
\newblock In {\em Proceedings of the 18th Annual SIGdial Meeting on Discourse
  and Dialogue, Saarbr{\"{u}}cken, Germany, August 15-17, 2017},  37--49.

\bibitem[\protect\citeauthoryear{Gasic \bgroup et al\mbox.\egroup
  }{2013}]{gasic2013pomdp}
Gasic, M.; Breslin, C.; Henderson, M.; Kim, D.; Szummer, M.; Thomson, B.;
  Tsiakoulis, P.; and Young, S.
\newblock 2013.
\newblock Pomdp-based dialogue manager adaptation to extended domains.
\newblock In {\em Proceedings of the SIGDIAL 2013 Conference},  214--222.

\bibitem[\protect\citeauthoryear{Jang, Gu, and Poole}{2016}]{JGP16}
Jang, E.; Gu, S.; and Poole, B.
\newblock 2016.
\newblock Categorical reparameterization with {G}umbel-softmax.
\newblock In {\em Proceedings of the International Conference on Learning
  Representations}.

\bibitem[\protect\citeauthoryear{Kingma and Ba}{2014}]{kin14}
Kingma, D.~P., and Ba, J.~L.
\newblock 2014.
\newblock Adam: A method for stochastic optimization.
\newblock In {\em Proc. 3rd Int. Conf. Learn. Representations}.

\bibitem[\protect\citeauthoryear{Levin, Pieraccini, and
  Eckert}{2000}]{levin2000stochastic}
Levin, E.; Pieraccini, R.; and Eckert, W.
\newblock 2000.
\newblock A stochastic model of human-machine interaction for learning dialog
  strategies.
\newblock {\em IEEE Transactions on Speech and Audio Processing} 8(1):11--23.

\bibitem[\protect\citeauthoryear{Li \bgroup et al\mbox.\egroup }{2015}]{Li15}
Li, J.; Galley, M.; Brockett, C.; Gao, J.; and Dolan, B.
\newblock 2015.
\newblock A diversity-promoting objective function for neural conversation
  models.
\newblock In {\em Proceedings of the 2016 Conference of the North American
  Chapter of the Association for Computational Linguistics: Human Language
  Technologies},  110--119.

\bibitem[\protect\citeauthoryear{Li \bgroup et al\mbox.\egroup }{2016}]{Li16}
Li, J.; Monroe, W.; Ritter, A.; Galley, M.; Gao, J.; and Jurafsky, D.
\newblock 2016.
\newblock Deep reinforcement learning for dialogue generation.
\newblock In {\em Proceedings of the 2016 Conference on Empirical Methods in
  Natural Language Processing},  1192--1202.

\bibitem[\protect\citeauthoryear{Maddison, Mnih, and Teh}{2016}]{MMT16}
Maddison, C.~J.; Mnih, A.; and Teh, Y.~W.
\newblock 2016.
\newblock The concrete distribution: A continuous relaxation of discrete random
  variables.
\newblock In {\em Proceedings of the International Conference on Learning
  Representations}.

\bibitem[\protect\citeauthoryear{Papineni \bgroup et al\mbox.\egroup
  }{2002}]{Pap02}
Papineni, K.; Roukos, S.; Ward, T.; and Zhu, W.-J.
\newblock 2002.
\newblock Bleu: a method for automatic evaluation of machine translation.
\newblock In {\em Proceedings of the 40th annual meeting on association for
  computational linguistics},  311--318.
\newblock Association for Computational Linguistics.

\bibitem[\protect\citeauthoryear{Ritter, Cherry, and
  Dolan}{2011}]{ritter2011data}
Ritter, A.; Cherry, C.; and Dolan, W.~B.
\newblock 2011.
\newblock Data-driven response generation in social media.
\newblock In {\em Proceedings of the conference on empirical methods in natural
  language processing},  583--593.
\newblock Association for Computational Linguistics.

\bibitem[\protect\citeauthoryear{Serban \bgroup et al\mbox.\egroup
  }{2016}]{Ser16}
Serban, I.~V.; Sordoni, A.; Bengio, Y.; Courville, A.~C.; and Pineau, J.
\newblock 2016.
\newblock Building end-to-end dialogue systems using generative hierarchical
  neural network models.
\newblock In {\em AAAI}, volume~16,  3776--3784.

\bibitem[\protect\citeauthoryear{Serban \bgroup et al\mbox.\egroup
  }{2017}]{Ser17}
Serban, I.~V.; Sordoni, A.; Lowe, R.; Charlin, L.; Pineau, J.; Courville,
  A.~C.; and Bengio, Y.
\newblock 2017.
\newblock A hierarchical latent variable encoder-decoder model for generating
  dialogues.
\newblock In {\em AAAI},  3295--3301.

\bibitem[\protect\citeauthoryear{Sordoni \bgroup et al\mbox.\egroup
  }{2015}]{Sor15}
Sordoni, A.; Galley, M.; Auli, M.; Brockett, C.; Ji, Y.; Mitchell, M.; Nie,
  J.-Y.; Gao, J.; and Dolan, B.
\newblock 2015.
\newblock A neural network approach to context-sensitive generation of
  conversational responses.
\newblock In {\em Proceedings of the 2015 Conference of the North American
  Chapter of the Association for Computational Linguistics: Human Language
  Technologies},  196--205.

\bibitem[\protect\citeauthoryear{Sutskever, Vinyals, and Le}{2014}]{Sut14}
Sutskever, I.; Vinyals, O.; and Le, Q.~V.
\newblock 2014.
\newblock Sequence to sequence learning with neural networks.
\newblock In {\em Advances in neural information processing systems},
  3104--3112.

\bibitem[\protect\citeauthoryear{van~den Oord, Vinyals, and
  others}{2017}]{Oor17}
van~den Oord, A.; Vinyals, O.; et~al.
\newblock 2017.
\newblock Neural discrete representation learning.
\newblock In {\em Advances in Neural Information Processing Systems},
  6306--6315.

\bibitem[\protect\citeauthoryear{Vinyals and Le}{2015}]{VL15}
Vinyals, O., and Le, Q.
\newblock 2015.
\newblock A neural conversational model.
\newblock {\em Proceedings of the International Conference on Machine Learning,
  Deep Learning Workshop.}

\bibitem[\protect\citeauthoryear{Walker, Prasad, and
  Stent}{2003}]{walker2003trainable}
Walker, M.; Prasad, R.; and Stent, A.
\newblock 2003.
\newblock A trainable generator for recommendations in multimodal dialog.
\newblock In {\em Eighth European Conference on Speech Communication and
  Technology}.

\bibitem[\protect\citeauthoryear{Wen \bgroup et al\mbox.\egroup
  }{2016}]{camrest}
Wen, T.-H.; Gasic, M.; Mrk\v{s}i\'{c}, N.; Rojas~Barahona, L.~M.; Su, P.-H.;
  Ultes, S.; Vandyke, D.; and Young, S.
\newblock 2016.
\newblock Conditional generation and snapshot learning in neural dialogue
  systems.
\newblock In {\em EMNLP},  2153--2162.
\newblock Austin, Texas: ACL.

\bibitem[\protect\citeauthoryear{Wen \bgroup et al\mbox.\egroup
  }{2017a}]{wen2017network}
Wen, T.; Vandyke, D.; Mrk{\v{s}}{\'\i}c, N.; Ga{\v{s}}{\'\i}c, M.;
  Rojas-Barahona, L.; Su, P.; Ultes, S.; and Young, S.
\newblock 2017a.
\newblock A network-based end-to-end trainable task-oriented dialogue system.
\newblock In {\em 15th Conference of the European Chapter of the Association
  for Computational Linguistics, EACL 2017-Proceedings of Conference},
  volume~1,  438--449.

\bibitem[\protect\citeauthoryear{Wen \bgroup et al\mbox.\egroup
  }{2017b}]{Wen17}
Wen, T.-H.; Miao, Y.; Blunsom, P.; and Young, S.
\newblock 2017b.
\newblock Latent intention dialogue models.
\newblock In {\em International Conference on Machine Learning},  3732--3741.

\bibitem[\protect\citeauthoryear{Williams and Young}{2007}]{Wil07}
Williams, J.~D., and Young, S.
\newblock 2007.
\newblock Partially observable {M}arkov decision processes for spoken dialog
  systems.
\newblock {\em Computer Speech \& Language} 21(2):393--422.

\bibitem[\protect\citeauthoryear{Williams}{1992}]{Wil92}
Williams, R.~J.
\newblock 1992.
\newblock Simple statistical gradient-following algorithms for connectionist
  reinforcement learning.
\newblock In {\em Reinforcement Learning}. Springer.
\newblock  5--32.

\bibitem[\protect\citeauthoryear{Zhao, Lee, and Eskenazi}{2018}]{ZLE18}
Zhao, T.; Lee, K.; and Eskenazi, M.
\newblock 2018.
\newblock Unsupervised discrete sentence representation learning for
  interpretable neural dialog generation.
\newblock In {\em Proceedings of the 56th Annual Meeting of the Association for
  Computational Linguistics (Volume 1: Long Papers)},  1098--1107.
\newblock Association for Computational Linguistics.

\end{thebibliography}
\bibliographystyle{aaai}
\end{document}